\pdfoutput=1

\documentclass{article} 
\usepackage{nips13submit_e}
\usepackage{times}
\usepackage{hyperref}
\usepackage{url}
\usepackage{graphicx}
\usepackage{subfigure}
\usepackage{amsmath,amssymb, amsthm}

\DeclareMathOperator*{\Diag}{Diag}

\newcommand{\R}{\mathbf{R}}
\newcommand{\ff}{\phi}
\newcommand{\Prob}{\mathbf{Prob}}
\newcommand{\E}{\mathbf{E}}

\newcommand{\hatS}{\hat S}
\newcommand{\ve}[2]{\la #1, #2\ra}
\newcommand{\eqdef}{:=}

\usepackage{algorithmic}
\usepackage{algorithm}
\let\la=\langle
\let\ra=\rangle
\theoremstyle{plain}
\newtheorem{theorem}{Theorem}

\theoremstyle{definition}

\newtheorem{assumption}[theorem]{Assumption}

\newtheorem{example}[theorem]{Example}

\title{On Optimal Probabilities in \\Stochastic Coordinate Descent Methods}

\author{
Peter Richt\'{a}rik and Martin Tak\'{a}\v{c}  \\ 
University of Edinburgh, United Kingdom\\
October 11, 2013
}

\nipsfinalcopy 

\begin{document}

\maketitle

\begin{abstract}
We propose and analyze a new parallel coordinate descent method---`NSync---in which at each iteration a random subset of coordinates is updated, in parallel, allowing for the subsets to be chosen \emph{non-uniformly.} We derive convergence rates under a strong convexity assumption, and comment on how to assign probabilities to the sets to optimize the bound. The complexity and practical performance of the method can outperform its uniform variant by an order of magnitude. Surprisingly, the strategy of updating a single randomly selected coordinate per iteration---with optimal probabilities---may require less iterations, both in theory and practice, than the strategy of updating all coordinates at every iteration.
\end{abstract}

\section{Introduction}

In this work we consider the optimization problem
\begin{equation}\label{eq:P}
 \min_{x\in\R^n} \ff(x),
\end{equation}
where $\ff$ is strongly convex and smooth.  We propose a new algorithm, and call it `NSync (Nonuniform SYNchronous Coordinate descent).

\begin{algorithm}[h!]
\caption{(`NSync)}
\begin{algorithmic} \label{algorithm:RCDM-smooth}
\STATE \textbf{Input:} Initial point $x^0 \in \R^n$, subset probabilities $\{p_S\}$ and stepsize parameters $w_1, \dots, w_n>0$
\FOR {$k=0,1,2,\dots$}
 \STATE Select a random set of coordinates $\hat{S}\subseteq \{1,\dots,n\}$ such that $\Prob(\hat{S}=S) = p_S$
 \STATE Updated selected coordinates: $x^{k+1} = x^k - \sum_{i\in \hat{S}} \frac1{w_i}\nabla_i \ff(x^k) e^i$
\ENDFOR
\end{algorithmic}
\end{algorithm}

In `NSync, we first assign a probability $p_S\geq 0$ to every subset $S$ of $[n]\eqdef \{1,\dots,n\}$, with $\sum_{S} p_S = 1$, and pick stepsize parameters $w_i>0$, $i=1,2,\dots,n$. At every iteration, a random set $\hat{S}$ is generated, independently from previous iterations, following the law $\Prob(\hat{S}=S)=p_S$, and then coordinates $i \in \hat{S}$ are updated in parallel by moving in the direction of the negative partial derivative with stepsize $1/w_i$. The updates are synchronized: no processor/thread is allowed to proceed before all updates are applied, generating the new iterate $x^{k+1}$. We specifically study samplings $\hat{S}$ which are \emph{non-uniform} in the sense that $p_i\eqdef \Prob(i \in \hat{S})= \sum_{S: i\in S} p_S$ is allowed to vary with $i$. By $\nabla_i \ff(x)$ we mean $\ve{\nabla \ff(x)}{e^i}$, where $e^i\in \R^n$ is the $i$-th unit coordinate vector.

\paragraph{Literature.} Serial stochastic coordinate descent methods were proposed and analyzed in \cite{SDCA-2008,Nesterov:2010RCDM, RT:UCDC,TewariShalevShwartzJMLR2011}, and more recently in various settings in \cite{Necoara:Coupled,Jaggi-ICML2013-block-Frank-Wolfe, lu2013complexity, lu2013on-monotone, Stoch-dual-Coord-Ascent, Proximal-dual-Coord-Ascent, TRG:InexactCDM, LanBCD2013}. Parallel methods were considered in \cite{Bradley:PCD-paper, RT:PCDM, RT:TTD2011}, and more recently in \cite{minibatch-ICML2013,FR:SPCDM2013,TRB2013:DQA,Fer-ParallelAdaboost,Necoara:parallelCDM-MPC, SSS2013-accelerated,BoostingMomentum2013,PCDN_2013}. A memory distributed method scaling to big data problems was recently developed in \cite{RT:Hydra2013}.  A nonuniform coordinate descent method updating a single coordinate at a time was proposed in \cite{RT:UCDC}, and one updating two coordinates at a time in \cite{Necoara:Coupled}. To the best of our knowledge, `NSync is the first \emph{nonuniform parallel} coordinate descent
method.

\section{Analysis}

Our analysis of `NSync is based on two assumptions. The first assumption generalizes the ESO concept introduced in \cite{RT:PCDM} and later used in \cite{minibatch-ICML2013,TRB2013:DQA,FR:SPCDM2013,Fer-ParallelAdaboost,RT:Hydra2013} to \emph{nonuniform} samplings. The second assumption requires that $\ff$ be strongly convex.

\emph{Notation:} For $x,y,u \in \R^n$ we write $\|x\|_u^2 \eqdef \sum_i u_i x_i^2$, $\ve{x}{y}_u \eqdef \sum_{i=1}^n u_i y_i x_i$, $x \bullet y \eqdef (x_1 y_1, \dots, x_n y_n)$ and $u^{-1} \eqdef (1/u_1,\dots,1/u_n)$. For $S\subseteq [n]$ and $h\in \R^n$, let $h_{[S]} \eqdef \sum_{i\in S} h_i e^i$.

\begin{assumption}[Nonuniform ESO: Expected Separable Overapproximation]  \label{ass:ESO}
Assume $p=(p_1,\dots,p_n)^T>0$ and that for some positive vector $w\in \R^n$ and all $x,h \in \R^n$,
\begin{equation}\label{eq:ESO}
 \E[ \ff(x+h_{[\hatS]})] \leq  \ff(x) + \la \nabla \ff(x), h \ra_p + \tfrac{1}{2} \|h\|_{p \bullet w}^2.
\end{equation}
Inequalities of type \eqref{eq:ESO}, in the \emph{uniform} case ($p_i=p_j$ for all $i,j$), were studied in \cite{RT:PCDM, minibatch-ICML2013,FR:SPCDM2013,RT:Hydra2013}.

\begin{assumption}[Strong convexity]  \label{ass:strong_convexity} We assume that $\ff$ is $\gamma$-strongly convex  with respect to the norm $\|\cdot\|_{v}$, where $v=(v_1,\dots,v_n)^T>0$ and $\gamma>0$. That is, we require that
for all $x,h \in \R^n$,
\begin{equation}\label{eq:asdf4eaf2wf}
\ff(x+h) \geq \ff(x) + \la \nabla \ff(x),h \ra + \tfrac{\gamma}2 \|h\|_{v}^2.
\end{equation}
\end{assumption}

\end{assumption}

We can now establish a bound on the number of iterations sufficient for `NSync to approximately solve \eqref{eq:P} with high probability.

\begin{theorem}\label{eq:convergence} Let Assumptions~\ref{ass:ESO} and \ref{ass:strong_convexity} be satisfied.
Choose $x^0 \in \R^n$, $0 < \epsilon < \ff(x^0)-\ff^*$ and $0< \rho < 1$, where $\ff^* \eqdef \min_x \ff(x)$. Let \begin{equation}\label{eq:Lambda}\Lambda \eqdef \max_{i} \tfrac{w_i}{p_i v_i}.\end{equation} If  $\{x^k\}$ are the random iterates generated by `NSync, then
\begin{equation}\label{eq:T}
 \textstyle{K \geq \tfrac{\Lambda}{\gamma}\log \left(\frac{\ff(x^0)-\ff^*}{\epsilon \rho}\right) \qquad \Rightarrow \qquad \Prob(\ff(x^{K})-\ff^* \leq \epsilon) \geq 1-\rho}.
\end{equation}
Moreover, we have the lower bound $\Lambda \geq (\sum_i \tfrac{w_i}{v_i})/\E[|\hat{S}|]$.
\end{theorem}
\vspace{-0.4cm}
\begin{proof}
We first claim that $\phi$ is $\mu$-strongly convex with respect to the norm $\|\cdot\|_{w\bullet p^{-1}}$, i.e.,
\begin{equation}\label{eq:7878787}\ff(x+h) \geq \ff(x) + \la \nabla \ff(x),h \ra + \tfrac{\mu}2 \|h\|_{w \bullet p^{-1}}^2,\end{equation}
where $\mu \eqdef \gamma/\Lambda$. Indeed, this follows by comparing \eqref{eq:asdf4eaf2wf} and \eqref{eq:7878787} in the light of \eqref{eq:Lambda}. Let $x^*$ be such that $\ff(x^*) = \ff^*$. Using \eqref{eq:7878787} with $h=x^*-x$,
\begin{equation}
 \ff^* - \ff(x)  \overset{\eqref{eq:7878787}}{\geq}    \min_{h' \in \R^n}  \la \nabla \ff(x), h'\ra + \tfrac{\mu}{2} \|h'\|_{ w \bullet p^{-1}}^2  =   -\tfrac1{2\mu}  \|\nabla \ff(x)\|^2_{p \bullet w^{-1}}.\label{eq:asdfjsafsa}
\end{equation}
Let $h^k \eqdef -(\Diag(w))^{-1}\nabla \ff(x^k)$. Then $x^{k+1}=x^k + (h^k)_{[\hat{S}]}$, and utilizing Assumption~\ref{ass:ESO}, we get 
\begin{gather}
\E[ \ff(x^{k+1}) \;|\; x^k ] =\E[ \ff(x^k + (h^k)_{[\hatS]}) ] \overset{\eqref{eq:ESO}}{\leq} \ff(x^k)
 + \la \nabla \ff(x^k), h^k \ra_p + \tfrac12 \|h^k\|_{p \bullet w}^2
\\
= \ff(x^k)
 - \tfrac12 \|  \nabla \ff(x^k)\|_{p \bullet w^{-1}}^2
 \overset{\eqref{eq:asdfjsafsa}}{\leq}
 \ff(x^k) - \mu (\ff(x^k)-\ff^*).
\end{gather}
Taking expectations in the last inequality and rearranging the terms, we obtain
$ \E[\ff(x^{k+1}) -\ff^*]\leq (1-\mu) \E[\ff(x^k)-\ff^*] \leq (1-\mu)^{k+1} (\ff(x^0)-\ff^*)$. Using this, Markov inequality, and the definition of $K$, we finally get $\Prob(\ff(x^K)-\ff^* \geq \epsilon) \leq \E[\ff(x^K)-\ff^*]/\epsilon \leq (1-\mu)^{K} (\ff(x^0)-\ff^*)/\epsilon \leq \rho$.
Let us now establish the last claim.
First, note that (see \cite[Sec 3.2]{RT:PCDM} for more results of this type),
\begin{equation}\label{eq:sum_of_pi}\textstyle{\sum_i p_i = \sum_i \sum_{S: i \in S}p_S = \sum_S \sum_{i : i \in S} p_S = \sum_S p_S |S|  = \E[|\hat{S}|].}\end{equation}
Letting $\Delta\eqdef \{p'\in \R^n : p'\geq 0, \sum_i p_i' = \E[|\hat{S}|]\}$, we have
\[\Lambda \overset{\eqref{eq:Lambda}+\eqref{eq:sum_of_pi}}{\geq} \min_{p' \in \Delta} \max_i \tfrac{w_i}{p_i' v_i}  = \tfrac{1}{\E[|\hat{S}|]}\sum_i \tfrac{w_i}{v_i},\]
where the last equality follows since optimal  $p_i'$ is proportional to $w_i/v_i$.
\end{proof}

Theorem~\ref{eq:convergence} is generic in the sense that we do not say when Assumptions 1 and 2 are satisfied, how should one go about to choose the stepsizes $w$ and probabilities $\{p_S\}$. In the next section we address these issues. On the other hand, this abstract setting allowed us to write a brief complexity proof.

\textbf{Change of variables.} Consider the change of variables $y=\Diag(d) x$, where $d>0$. Defining $\ff^d(y)\eqdef \ff(x)$, we get
$\nabla \ff^d(y) = (\Diag(d))^{-1}\nabla \ff(x)$. It can be seen that \eqref{eq:ESO}, \eqref{eq:asdf4eaf2wf} can equivalently be written in terms of $\ff^d$, with $w$ replaced by $w^d \eqdef w \bullet d^{-2}$ and $v$ replaced by $v^d \eqdef v \bullet d^{-2}$. By choosing $d_i=\sqrt{v_i}$, we obtain  $v^d_i=1$ for all $i$, recovering standard strong convexity. 

\section{Nonuniform samplings and ESO}

Consider now problem \eqref{eq:P} with $\ff$ of the form
\begin{equation}\label{eq:Pweighted}\textstyle{\ff(x) \eqdef f(x) + \tfrac{\gamma}{2}\|x\|_v^2,}\end{equation}
where $v>0$. Note that Assumption~\ref{ass:strong_convexity} is satisfied. We further make the following two assumptions.

\begin{assumption}[Smoothness]\label{ass:Lip}
$f$ has Lipschitz gradient with respect to the coordinates, with positive constants
$L_1,\dots,L_n$. That is, $  | \nabla_i f(x) -\nabla_i f(x+t e_i)|  \leq L_i |t|$ for all $ x \in \R^n$ and $t\in \R$.
\end{assumption}

\begin{assumption}[Partial separability]\label{ass:partial_sep}  $f(x) = \sum_{J \in \mathcal{J}} f_J (x)$, where $\mathcal{J}$ is a finite collection of nonempty subsets of $[n]$ and $f_J$ are differentiable convex functions such that $f_J$ depends on coordinates $i\in J$ only. Let $\omega \eqdef \max_{J} |J|$. We say that $f$ is \emph{separable of degree $\omega $.}
\end{assumption}

\emph{Uniform} parallel coordinate descent methods for regularized problems with $f$ of the above structure were analyzed in \cite{RT:PCDM}.

\begin{example}\label{ex:example} Let $f(x)=\tfrac{1}{2}\|Ax-b\|_2^2$, where $A \in \R^{m \times n}$. Then $L_i = \|A_{:i}\|_2^2$ and $f(x) = \tfrac{1}{2}\sum_{j=1}^m (A_{j:} x - b_j)^2$, whence $\omega$ is the maximum \# of nonzeros in a row of $A$.
\end{example}

\paragraph{Nonuniform sampling.} Instead of considering the general case of arbitrary $p_S$ assigned to all subsets of $[n]$, here we consider a special kind of sampling having two advantages: i) sets can be generated easily, ii) it leads to larger stepsizes $1/w_i$ and hence improved convergence rate. Fix $\tau \in [n]$ and  $c\geq 1$ and let  $S_1,\dots,S_c$ be a collection of (possibly overlapping) subsets of $[n]$ such that $|S_j| \geq \tau$ for all $i$ and  $\cup_{j=1}^c S_j = [n]$. Moreover, let  $q= (q_1,\dots,q_c)> 0$ be a probability vector. Let $\hat{S}_j$ be $\tau$-nice sampling from  $S_j$; that is, $\hat{S}_j$ picks subsets of $S_j$ having cardinality $\tau$, uniformly at random. We assume these samplings are independent. Now, $\hat{S}$ is generated as follows. We first pick $j\in \{1,\dots,c\}$ with probability $q_j$, and then draw $\hat{S}_j$. Note that we do not need to compute the quantities $p_S$, $S\subseteq [n]$, to execute `NSync. In fact, it is much easier to implement the sampling via the two-tier procedure explained above. Sampling $\hat{S}$ is a nonuniform variant of the $\tau$-nice sampling studied in \cite{RT:PCDM}, which here arises as a special case for $c=1$. Note that
\begin{equation}\label{eq:p_i} \textstyle{p_i   = \sum_{j=1}^c q_j\frac{\tau }{|S_j|} \delta_{ij} >0, \quad i \in [n],}\end{equation}
where $\delta_{ij}=1$ if $i \in S_j$, and $0$ otherwise.
\begin{theorem}\label{thm:ESOFORCOmposite}
Let Assumptions~\ref{ass:Lip} and \ref{ass:partial_sep} be satisfied, and let $\hat{S}$ be the sampling described above. Then Assumption~\ref{ass:ESO} is satisfied with $p$ given by \eqref{eq:p_i} and any $w=(w_1,\dots,w_n)^T$ for which
\begin{equation}\label{eq:w_i}
\textstyle{w_i  \geq w_i^*  \eqdef \frac{L_i+v_i}{p_i}\sum_{j=1}^c  q_j\frac{\tau }{|S_j|} \delta_{ij} \left(1+\frac{(\tau-1)(\omega_j-1)}{\max\{1,|S_j|-1\}}\right) , \qquad i \in [n],}
\end{equation}
where $\omega_j \eqdef \max_{J \in \mathcal{J}} |J \cap S_j| \leq \omega$.
\end{theorem}
\vspace{-0.4cm}
\begin{proof}
Since $f$ is separable of degree $\omega$, so is $\phi$ (because $\frac{1}{2}\|x\|_v^2$ is separable). Now,
\begin{gather}
\textstyle{\E[\ff(x+h_{[\hatS]})]}
= \textstyle{\E[\E[\ff(x+h_{[\hatS_j]}) \;|\; j ]]  =\sum_{j=1}^c   q_j \E[\ff(x+h_{[\hatS_j]})]}
\\
\textstyle{\leq \sum_{j=1}^c
  q_j \left\{ f(x) +
   \tfrac{\tau}{|S_j|}  \left( \la \nabla f(x), h_{[S_j]}\ra
    + \tfrac{1}{2}\left(
         1+\tfrac{  (\tau-1)(\omega_j-1)}{\max\{1,|S_j|-1\}}
         \right) \|h_{[S_j]}\|_{L+v}^2
   \right)  \right\},}
\end{gather}
where the last inequality follows from the ESO for $\tau$-nice samplings established in \cite[Theorem 15]{RT:PCDM}. The claim now follows by comparing the above expression and \eqref{eq:ESO}.
\end{proof}


\section{Optimal probabilities}

Observe that formula \eqref{eq:w_i} can be used to \emph{design} a sampling (characterized by the sets $S_j$ and probabilities $q_j$) that \emph{minimizes} $\Lambda$, which in view of Theorem~\ref{eq:convergence} \emph{optimizes the convergence rate} of the method.

\textbf{Serial setting.} Consider the serial version of `NSync ($\Prob(|\hat{S}|=1)=1$). We can model this via $c=n$, with $S_i =\{i\}$ and $p_i=q_i$ for all $i \in [n]$. In this case, using \eqref{eq:p_i} and \eqref{eq:w_i}, we get $w_i = w_i^* =L_i + v_i$. Minimizing $\Lambda$ in  \eqref{eq:Lambda} over the probability vector $p$ gives the \emph{optimal probabilities} (we refer to this as the \emph{optimal serial} method) and \emph{optimal complexity}
\begin{equation}\label{eq:p-optimal}p_i^*  = \tfrac{(L_i+v_i)/v_i}{\sum_j (L_j+v_j)/v_j }, \quad i \in [n], \qquad \qquad  \textstyle{\Lambda_{OS} = \sum_i \tfrac{L_i+v_i}{v_i} = n+ \sum_i \tfrac{L_i}{v_i},}\end{equation}
respectively. Note that the \emph{uniform sampling}, $p_i=1/n$ for all $i$, leads to $\Lambda_{US} \eqdef n + n \max_{j} L_j/v_j$ (we call this the \emph{uniform serial} method), which can be much larger than $\Lambda_{OS}$. Moreover, under the change of variables $y=\Diag(d)x$, the gradient of $f^d(y)\eqdef f(\Diag(d^{-1})y)$ has coordinate Lipschitz constants $L_i^d = L_i/d_i^2$, while the weights in \eqref{eq:Pweighted} change to $v_i^d = v_i/d_i^2$. Hence, the condition numbers $L_i/v_i$ can not be improved via such a change of variables. 

\textbf{Optimal serial method can be faster than the fully parallel method.} To model the fully parallel setting (i.e., the variant of `NSync updating \emph{all} coordinates at every iteration), we can set $c=1$ and $\tau=n$, which yields $\Lambda_{FP} = \omega + \omega\max_j L_j/v_j$. Since $\omega \leq n$, it is clear that $\Lambda_{US} \geq \Lambda_{FP}$. However, for large enough $\omega$ it will be the case that $\Lambda_{FP}\geq \Lambda_{OS}$, implying, surprisingly, that the optimal serial method can be faster than the fully parallel method.

\textbf{Parallel setting.} Fix $\tau$ and sets $S_j$, $j=1,2,\dots,c$, and  define $\theta \eqdef \max_j\left( 1+ \tfrac{(\tau-1)(\omega_j-1)}{\max\{1,|S_j|-1\}}\right)$. Consider running `NSync with  stepsizes $w_i = \theta(L_i+v_i)$ (note that $w_i \geq w_i^*$, so we are fine). From \eqref{eq:Lambda}, \eqref{eq:p_i}  and \eqref{eq:w_i} we see that the complexity of `NSync is determined by
\[\textstyle{\Lambda = \max_i \tfrac{w_i}{p_i v_i} = \tfrac{\theta}{\tau} \max_i \left(1+ \tfrac{L_i}{v_i}\right) \left(\sum_{j=1}^c q_j\frac{\delta_{ij}}{|S_j|}\right)^{-1}.}\]
The probability vector $q$ minimizing this quantity can be computed by solving a linear program with $c+1$ variables ($q_1,\dots,q_c,\alpha$), $2n$ linear inequality constraints  and a single linear equality constraint:
\[\textstyle{\max_{\alpha,q} \left\{\alpha \;  \text{ subject to }\; \alpha \leq (b^i)^T q  \text{ for all } i, \; q\geq 0, \; \sum_j q_j = 1 \right\},}\]
where $b^i \in \R^c$, $i\in [n]$, are given by $b^i_j = \tfrac{v_i}{(L_i+v_i)}\tfrac{\delta_{ij}}{|S_j|}$.

\section{Experiments}

We now conduct 2 preliminary small scale experiments to illustrate the theory; the results are depicted below. All experiments are with problems of the form \eqref{eq:Pweighted} with $f$ chosen as in Example~\ref{ex:example}.

{
 \centering
 \includegraphics[height=1.5in]{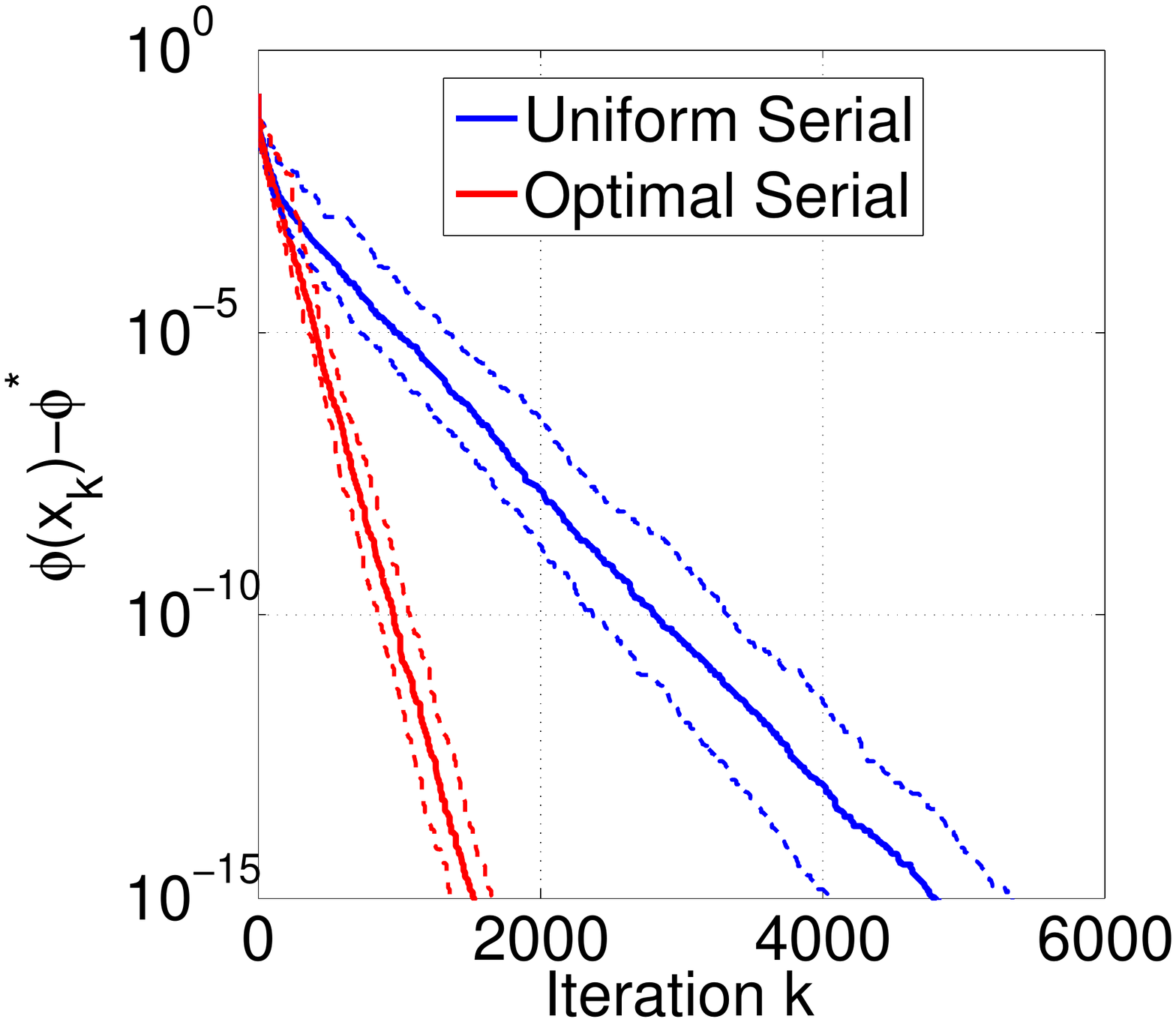}
 \includegraphics[height=1.5in]{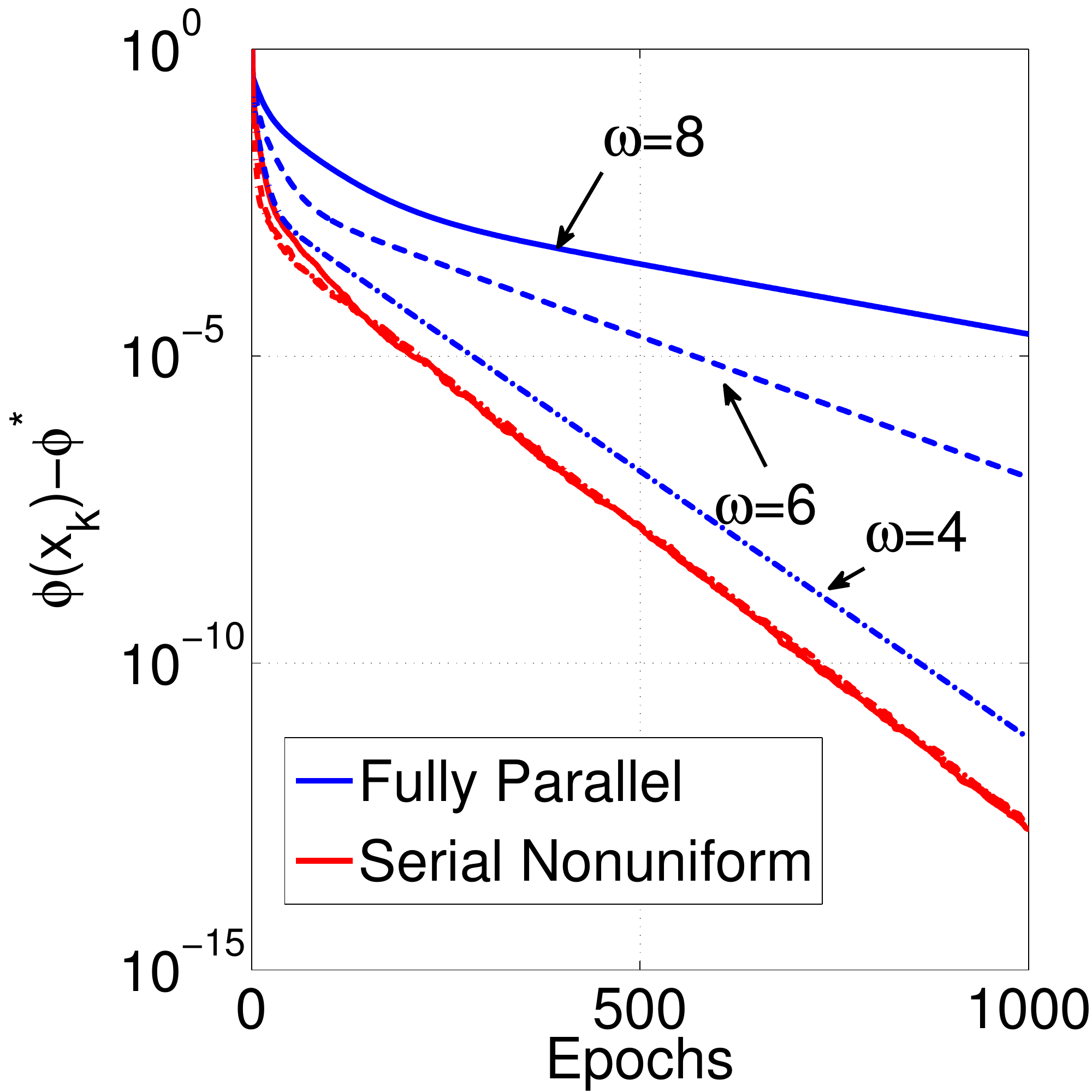}

}

In the \textbf{left plot} we chose $A\in \R^{2\times 30}$, $\gamma=1$, $v_1=0.05$, $v_i=1$ for $i\neq 1$ and $L_i=1$ for all $i$. We compare the US  method ($p_i = 1/n$, blue) with the  OS  method ($p_i$ given by \eqref{eq:p-optimal}, red). The dashed lines show 95\% confidence intervals (we run the methods 100 times, the line in the middle is the average behavior). While OS can be faster, it is sensitive to over/under-estimation of the constants $L_i,v_i$. In the \textbf{right plot} we show that a nonuniform serial (NS) method can be faster than the fully parallel (FP) variant (we have chosen $m=8$, $n=10$ and 3 values of $\omega$). On the horizontal axis we display the number of epochs, where 1 epoch corresponds to updating $n$ coordinates (for FP this is a  single iteration, whereas for NS it corresponds to $n$ iterations).

%
%




\clearpage
\small{

 \bibliographystyle{plain} 
\bibliography{literature}

}

\end{document}